\documentclass[runningheads]{llncs}
\usepackage[T1]{fontenc}
\usepackage{graphicx,verbatim}
\usepackage{hyperref}
\usepackage{color}

\usepackage{amsmath}
\usepackage{multirow}

\begin{document}
\title{Sparse Autoencoders for Interpretable Medical Image Representation Learning}
\titlerunning{SAEs for Interpretable Medical Image Representation Learning}
%
\author{Philipp Wesp\inst{1,2} \and
Robbie Holland\inst{1,2} \and
Vasiliki Sideri-Lampretsa\inst{3,4} \and
Sergios Gatidis\inst{1,2}}
\authorrunning{P. Wesp et al.}
%

\institute{Stanford Center for Artificial Intelligence in Medicine and Imaging, Stanford University, Stanford, CA, USA \and
Department of Radiology, Stanford University, Stanford, CA, USA \and
Chair of AI in Healthcare and Medicine, TUM, Munich, Germany \and
TUM University Hospital, Munich, Germany}


  
\maketitle              
\begin{abstract}

Vision foundation models (FMs) achieve state-of-the-art performance in medical imaging.
However, they encode information in abstract latent representations that clinicians cannot interrogate or verify.
The goal of this study is to investigate Sparse Autoencoders (SAEs) for replacing opaque FM image representations with human-interpretable, sparse features.
We train SAEs on embeddings from BiomedParse (biomedical) and DINOv3 (general-purpose) using 909,873 CT and MRI 2D images from the TotalSegmentator dataset.
We find that learned sparse features:
(a) reconstruct original embeddings with high fidelity (R$^2$ up to 0.941) and recover up to 87.8\,\% of downstream performance using only 10 features (99.4\,\% dimensionality reduction),
(b) preserve semantic fidelity in image retrieval tasks,
(c) correspond to specific concepts that can be expressed in language using large language model (LLM)-based auto-interpretation,
(d) bridge clinical language and abstract latent representations in zero-shot language-driven image retrieval.
Our work indicates SAEs are a promising pathway towards interpretable, concept-driven medical vision systems.
Code repository: \url{https://github.com/pwesp/sail}.


\end{abstract}
\section{Introduction}

\begin{figure}[t]
    \centering
    \includegraphics[width=0.8\textwidth]{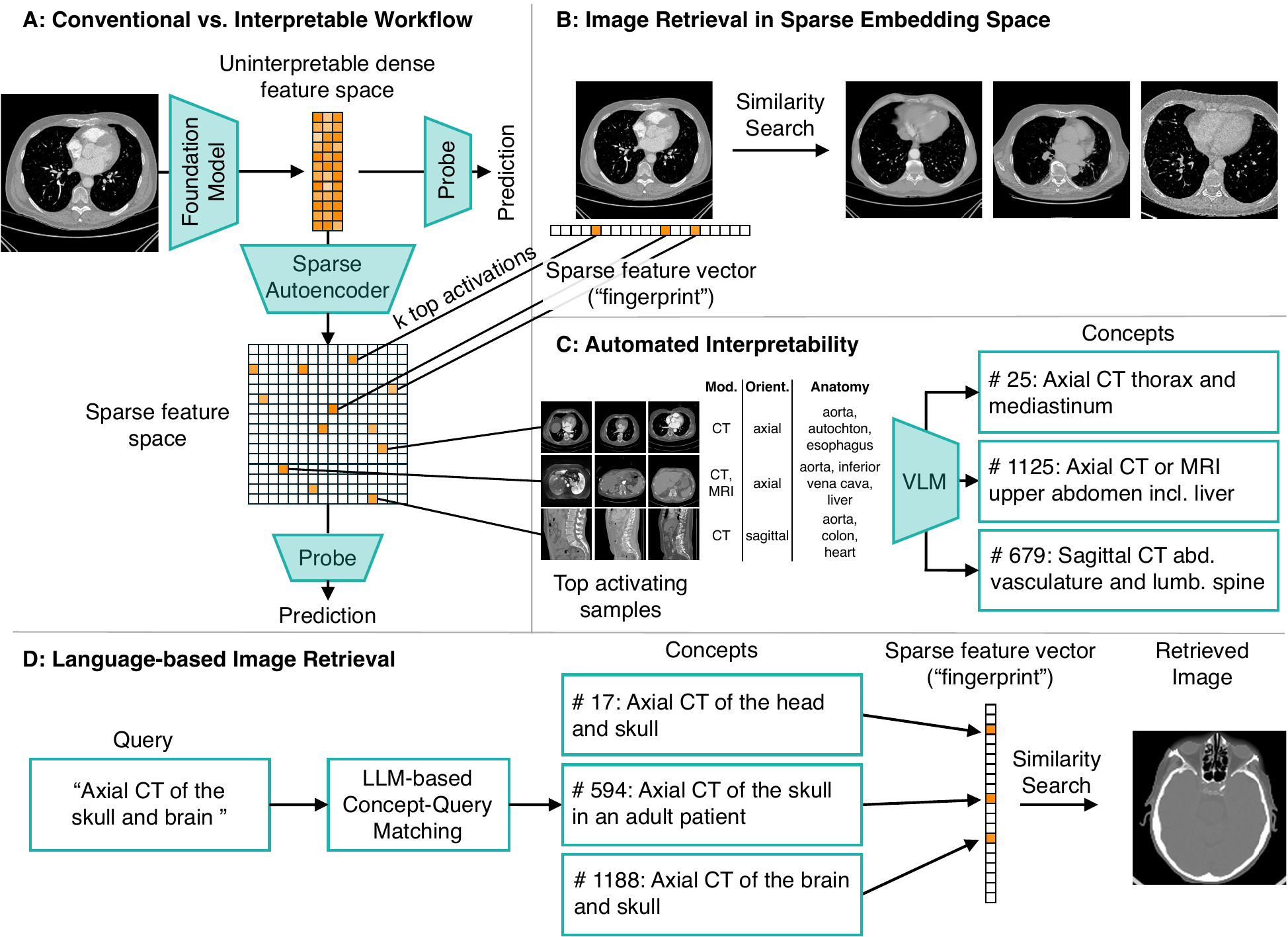}
    \caption{(A) A Sparse Autoencoder replaces opaque dense FM embeddings with a sparse feature space.
    (B) Sparse fingerprint retrieval matches images by cosine similarity over $k$ top-activated features.
    (C) A VLM generates a concept description for each feature from its top-activating images and metadata.
    (D) An LLM maps a clinical text query to matching feature concepts for zero-shot image retrieval.}
    \label{fig:graphical-abstract}
\end{figure}

Vision foundation models (FMs) achieve strong performance in medical imaging tasks such as segmentation, classification, and retrieval, but encode information in abstract, low-dimensional feature representations \cite{moor2023nature,paschali2025radiology}.
At the same time, clinical deployment demands interpretability: physicians must justify decisions, detect failure modes, and document reasoning, yet model internals remain inaccessible \cite{langlotz2019radiology}.
This creates a fundamental misalignment between abstract learned representations and the anatomical and clinical concepts that clinicians reason with.

Mechanistic interpretability aims to reverse-engineer model internals into human-understandable components.
Sparse Autoencoders (SAEs) \cite{bricken2023transformercircuitsthread,cunningham2023} are a leading approach, decomposing polysemantic activations in large language models (LLMs) into monosemantic features \cite{elhage2022transformercircuitsthread} that each correspond to a single coherent concept.
A recent study applied SAEs to chest radiograph embeddings, demonstrating that a small number of interpretable sparse features can represent clinically relevant visual concepts and support radiology report generation \cite{abdulaal2024}.
That study, however, was restricted to a single modality and a single FM architecture with paired text supervision for concept labelling, leaving open whether anatomical structure emerges in self-supervised models across CT, MRI, and diverse anatomical regions.
This raises a central question: do anatomical and clinical concepts emerge from self-supervised medical vision training without explicit labels, and can SAEs expose this structure consistently across architecturally distinct foundation models?
Because whole-body CT and MRI provide no paired text, we introduce a grounding procedure that scores VLM- and segmentation-derived concept labels against a null baseline to make monosemanticity measurable, together with a configuration-ranking framework that exposes the monosemanticity-performance trade-off.

To this end, we train Matryoshka SAEs \cite{bussmann2025} with BatchTopK sparsification \cite{bussmann2024} on frozen embeddings from BiomedParse \cite{zhao2025natmethods} (biomedical FM) and DINOv3 \cite{simeoni2025} (general-purpose FM), alongside a random-weight baseline to isolate learned representational structure from architectural effects, across 909,873 CT and MRI 2D images from the TotalSegmentator dataset \cite{wasserthal2023radiol_artif_intell,akincidantonoli2025radiology} (Fig. \ref{fig:graphical-abstract}).
We find that sparse features (a) faithfully reconstruct dense embeddings (R$^2$ up to 0.941) and recover 87.8\,\% of downstream performance with only 10 features, (b) preserve 97.7\,\% of dense retrieval quality with five-feature fingerprints, (c) correspond to monosemantic concepts verified by an independent large language model judge, and (d) enable zero-shot language-driven image retrieval bridging clinical text and medical image content.
These findings indicate that self-supervised vision FMs implicitly encode anatomy-aligned structure that SAEs can expose as language-describable sparse features, a step toward interpretable medical AI aligned with human language.

\section{Methods}

We train SAEs (the only optimized parameters) on frozen, precomputed embeddings from three vision FMs: BiomedParse \cite{zhao2025natmethods} (1536-dim, biomedical FM), DINOv3 \cite{simeoni2025} (1024-dim, general-purpose self-supervised ViT), and an untrained BiomedParse model with randomly initialised weights (1536-dim, random-weight baseline) to isolate learned representational structure from architectural effects.

\subsection{Sparse Autoencoder}

We adopt the Matryoshka SAE architecture \cite{bussmann2025} with $L=4$ nested dictionary levels of increasing size $[D_1, D_2, D_3, D_4]$.
A single shared linear encoder projects the input into $D_4$ pre-activation codes.
Level $\ell$ uses only the first $D_\ell$ codes as a prefix subset, so that early levels capture coarse structure and later levels refine it progressively.
A single shared decoder (encoder weights transposed, columns normalised to unit norm) reconstructs the input at each level by padding smaller-level activations with zeros.
During training, we apply BatchTopK sparsification \cite{bussmann2024}: $k$ features are active per sample on average across the batch, allowing flexible per-sample allocation unlike fixed per-sample TopK.
At inference, a JumpReLU \cite{rajamanoharan2024} threshold, estimated as a running average of the minimum kept activation during training, replaces BatchTopK.
The training objective is the mean squared error (MSE) between input and reconstruction, averaged across all $L$ levels, with no auxiliary sparsity or diversity penalties.

\paragraph{Monosemanticity scoring.}
To compare configurations, we score each feature as $M(f) = C(f)\,S(f)$.
Coherence $C(f) = \binom{n}{2}^{-1}\sum_{i<j} J(O_i, O_j)$ is the mean pairwise Jaccard similarity $J$ over the organ-presence sets $O_i$ of the $n=10$ top-activating samples of $f$.
Specificity $S(f) = 1 - H(p)/\log K$ is the normalized inverse Shannon entropy of the organ-count distribution $p$ over those samples, where $K=121$ is the number of organ labels.
To distinguish genuine structure from chance co-occurrence, we estimate a null baseline $\bar{C}_{\mathrm{null}}$ (mean of $C$ over 100 random 10-sample subsets), and call a feature above-null when $C(f) > \bar{C}_{\mathrm{null}}$.
The configuration-level score $M_{\mathrm{config}}$ is the mean $M(f)$ of the top-10 features per configuration.

\subsection{Interpretability Evaluation}

We evaluate the interpretability of learned sparse features through three complementary demonstrations.

\paragraph{Sparse Fingerprint Retrieval.}
We define a sparse fingerprint as the $k$ most activated features and their values per image, and retrieve similar images by cosine similarity over fingerprints.
Retrieval quality is measured as mean cosine similarity to the reference in the dense embedding space, with dense retrieval as the upper bound.

\paragraph{Automated Feature Interpretation.}
To assess whether individual features encode interpretable and consistent concepts, we greedily select the 5 most dissimilar samples from the top-20 activating images for the top-100 most monosemantic ($M$ score) features using cosine similarity.
We then prompt the vision language model (VLM) MedGemma 27B \cite{sellergren2025} to generate a natural-language concept description from their images and metadata (modality, orientation, anatomy, demographics) \cite{hernandez2022,bills2023openai}.
A VLM judge (separate MedGemma 27B) then receives the same images and five candidate descriptions and must identify the correct one.
The four distractors are sampled uniformly from the concept descriptions of other not-always-active features with more than 10 activations, which excludes trivial and degenerate cases.
Generation and judging run as independent inference calls with different prompts, so the task measures generic vision-language matching rather than self-recognition.
The rank of the true concept ($1 = \text{best}$, $5 = \text{worst}$) quantifies interpretability, where chance corresponds to a mean rank of 3.

\paragraph{Language-Driven Image Retrieval.}
An LLM identifies feature descriptions that match a clinical text query, assembling a sparse fingerprint from their mean activations for cosine retrieval without a reference image.
This zero-shot procedure demonstrates that sparse feature concepts can bridge human language and medical image content.

\section{Experiments \& Results}
\label{sec:results}

\begin{figure}[t]
    \centering
    \includegraphics[width=\textwidth]{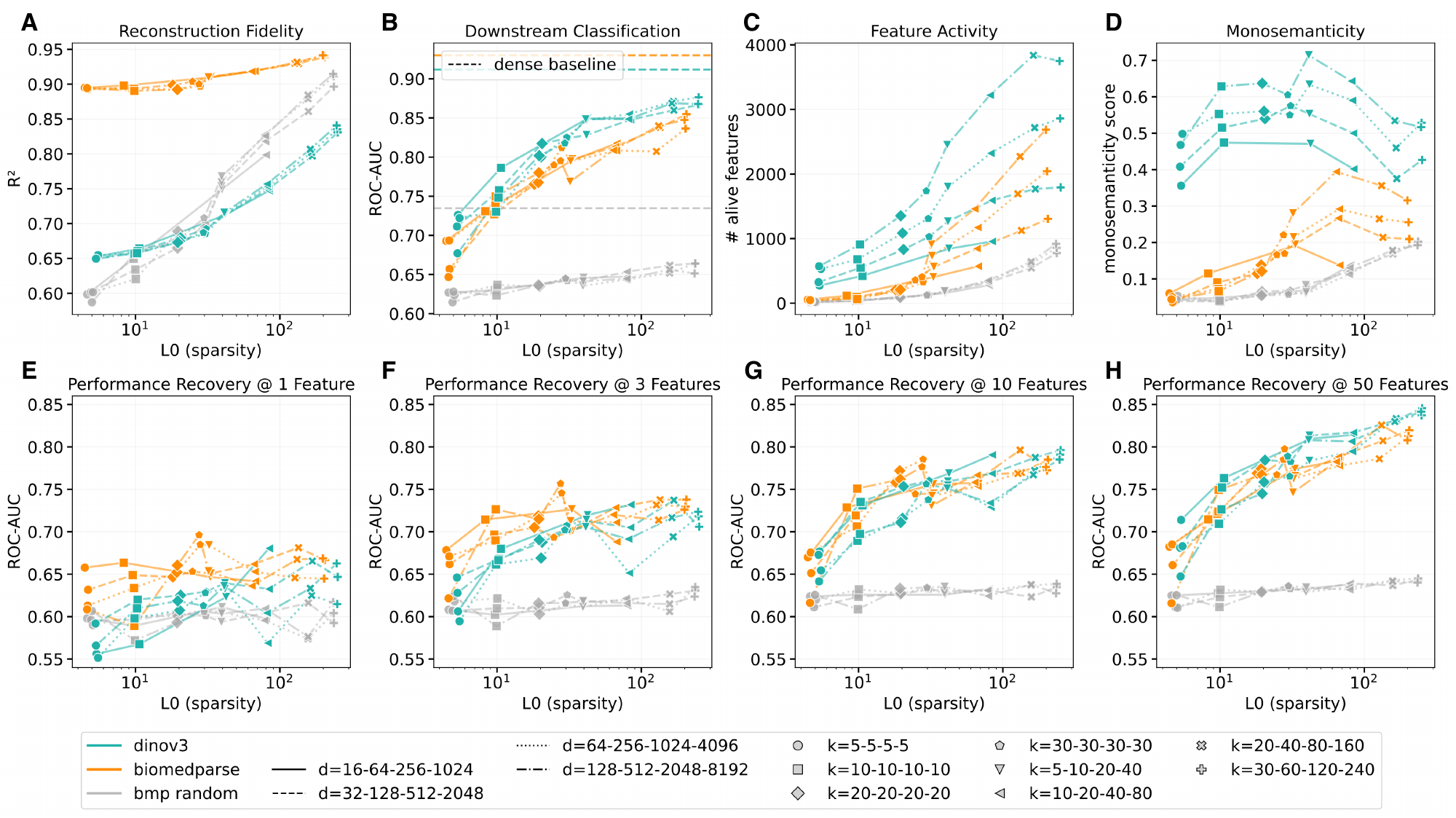}
    \caption{SAE quality and performance recovery across 32 configurations per FM (DINOv3: blue, BiomedParse: orange, random baseline: grey).
    (A--D) Reconstruction fidelity (R$^2$), downstream ROC-AUC, alive features, and monosemanticity score vs. L0 sparsity.
    (E--H) Performance recovery using only the top-$N$ features ($N=1,3,10,50$).}
    \label{fig:sae-overview}
\end{figure}

We evaluate Matryoshka SAEs on BiomedParse and DINOv3 embeddings from the TotalSegmentator dataset: 1,844 scans (1,228 CT, 616 MRI) from 10 institutions, yielding 909,873 2D images with 138 per-image metadata fields spanning anatomy presence, imaging parameters and demographics.
Scans from three institutions are withheld entirely as a test set (14.1\,\% of images), and the remaining scans are split 80/20 into train and validation sets stratified by modality, age group, and sex, yielding 68.6\,\% and 17.3\,\% of images respectively.
CT volumes use one of 38 study-type-dependent Hounsfield-unit windows (e.g. thorax $-200/1400$, abdomen $50/1000$), and MRI volumes are resampled to isotropic in-plane spacing and clipped to the 1st--99th intensity percentile.
Both are rescaled to $[0,255]$, center-cropped, resized to $1024\times1024$, replicated to three channels, and ImageNet-normalized.
SAEs are optimized with Adam \cite{kingma20153rdint_conf_learn_represent_iclr} (lr$=10^{-4}$, cosine annealing to $10^{-6}$, 100 epochs, batch size 2048) across 32 configurations per FM: 4 dictionary size families ($[16,64,256,1024]$ to $[128,512,2048,8192]$) and 8 sparsity patterns (4 fixed, 4 progressive $K$).
Baselines are a dense embedding upper bound and a random-weight BiomedParse model isolating learned structure from architectural effects.

\subsection{SAE Quality}
\label{subsec:sae-quality}

\paragraph{Latent space reconstruction (R$^2$).}

Figure \ref{fig:sae-overview} shows reconstruction quality, downstream performance, and alive feature counts across all 32 configurations per FM.
R$^2$ ranges from 0.890 to 0.941 for BiomedParse and from 0.649 to 0.841 for DINOv3.

\paragraph{Downstream performance (ROC-AUC).}

Dense embedding baselines achieve ROC-AUC of 0.907 (BiomedParse) and 0.912 (DINOv3) across anatomical classification tasks.
Optimal sparse configurations recover 90.2\% and 93.0\% of dense performance, respectively.
The random-weight baseline reaches only 0.606--0.651 AUC despite a comparable R$^2$ range (0.587--0.915), confirming that downstream utility reflects learned representational structure, not architectural capacity alone.
This dissociation shows that reconstruction fidelity is an insufficient proxy for semantic utility, since a sparse code can faithfully reconstruct a random embedding space while encoding no semantically meaningful structure.
Conversely, DINOv3's lower R$^2$ (0.649--0.841) relative to BiomedParse (0.890--0.941) coexists with higher downstream AUC, indicating that task-relevant structure can be preserved under approximate reconstruction.

\begin{table}[t]
    \centering
    \caption{Top-3 SAE configurations per FM ranked by combined monosemanticity and performance recovery score (32 configurations each). Bold: selected optimal configuration. \#$_{\text{Mono}}$/$_{\text{Perf}}$/$_{\text{Comb}}$: monosemanticity/performance/combined rank.}
    \label{tab:final-ranking}
    \scriptsize
    \setlength{\tabcolsep}{6pt}
    \begin{tabular}{lllcc|c}
    \hline
    \textbf{Model} & \textbf{Dict Sizes} & \textbf{Top-K Values} & \textbf{\#$_{\text{Mono}}$} & \textbf{\#$_{\text{Perf}}$} & \textbf{\#$_{\text{Comb}}$} \\
    \hline
    \multirow{3}{*}{BiomedParse} & \textbf{128, 512, 2048, 8192} & \textbf{20, 40, 80, 160} & 2 & \textbf{3} & 1 \\
    & 128, 512, 2048, 8192 & 30, 60, 120, 240 & 3 & 6 & 2 \\
    & 128, 512, 2048, 8192 & 10, 20, 40, 80 & 1 & 12 & 3 \\
    \hline
    \multirow{3}{*}{DINOv3} & \textbf{128, 512, 2048, 8192} & \textbf{5, 10, 20, 40} & 1 & \textbf{11} & 1 \\
    & 64, 256, 1024, 4096 & 5, 10, 20, 40 & 4 & 10 & 2 \\
    & 128, 512, 2048, 8192 & 10, 20, 40, 80 & 2 & 14 & 3 \\
    \hline
    \end{tabular}
\end{table}

\subsection{SAE Configuration Ranking}
\label{subsec:sae-configuration-ranking}

\paragraph{Monosemanticity \& performance recovery.}
We quantify the competing properties of monosemanticity and performance recovery \cite{karvonen2025} across all configurations and select an optimal configuration per model based on a combined score.
Figure \ref{fig:sae-overview} shows $M_{\mathrm{config}}$ and performance recovery.
DINOv3 achieves substantially higher monosemanticity (0.356--0.714) than BiomedParse (0.036--0.394), despite BiomedParse's domain-specific pretraining.
The random-weight baseline (0.038--0.202) falls well below both learned models, confirming that monosemanticity reflects learned representational structure rather than architectural capacity.
Within the optimal SAEs, the Matryoshka levels realize their intended coarse-to-fine organization, as median feature activation frequency falls from roughly 40,000 (BiomedParse) and 17,000 (DINOv3) images at level 0 to 1,200 and 152 at level 3.
The deepest level also holds the majority of coherent (above-null) features, 2,610/3,682 (BiomedParse) and 2,361/3,577 (DINOv3), concentrating organ-specific concepts in deep levels.
With $N=10$ features, BiomedParse recovers 87.8\% and DINOv3 recovers 82.4\% of dense ROC-AUC, with performance gains diminishing above $N=10$.

\paragraph{Configuration ranking.}

Table \ref{tab:final-ranking} ranks the top-3 configurations per model by combined monosemanticity and performance recovery score.
Progressive Top-K patterns with the largest dictionary family $[128,512,2048,8192]$ dominate the BiomedParse and DINOv3 rankings.
BiomedParse's optimal configuration ($K=[20,40,80,160]$) achieves competitive scores on both dimensions (monosemanticity rank 2, performance rank 3).
DINOv3's optimal ($K=[5,10,20,40]$) leads in monosemanticity (rank 1) but ranks 11th in performance recovery, exemplifying the monosemanticity-performance trade-off inherent to sparser representations.

\begin{figure}[t]
    \centering
    \includegraphics[width=0.8\textwidth]{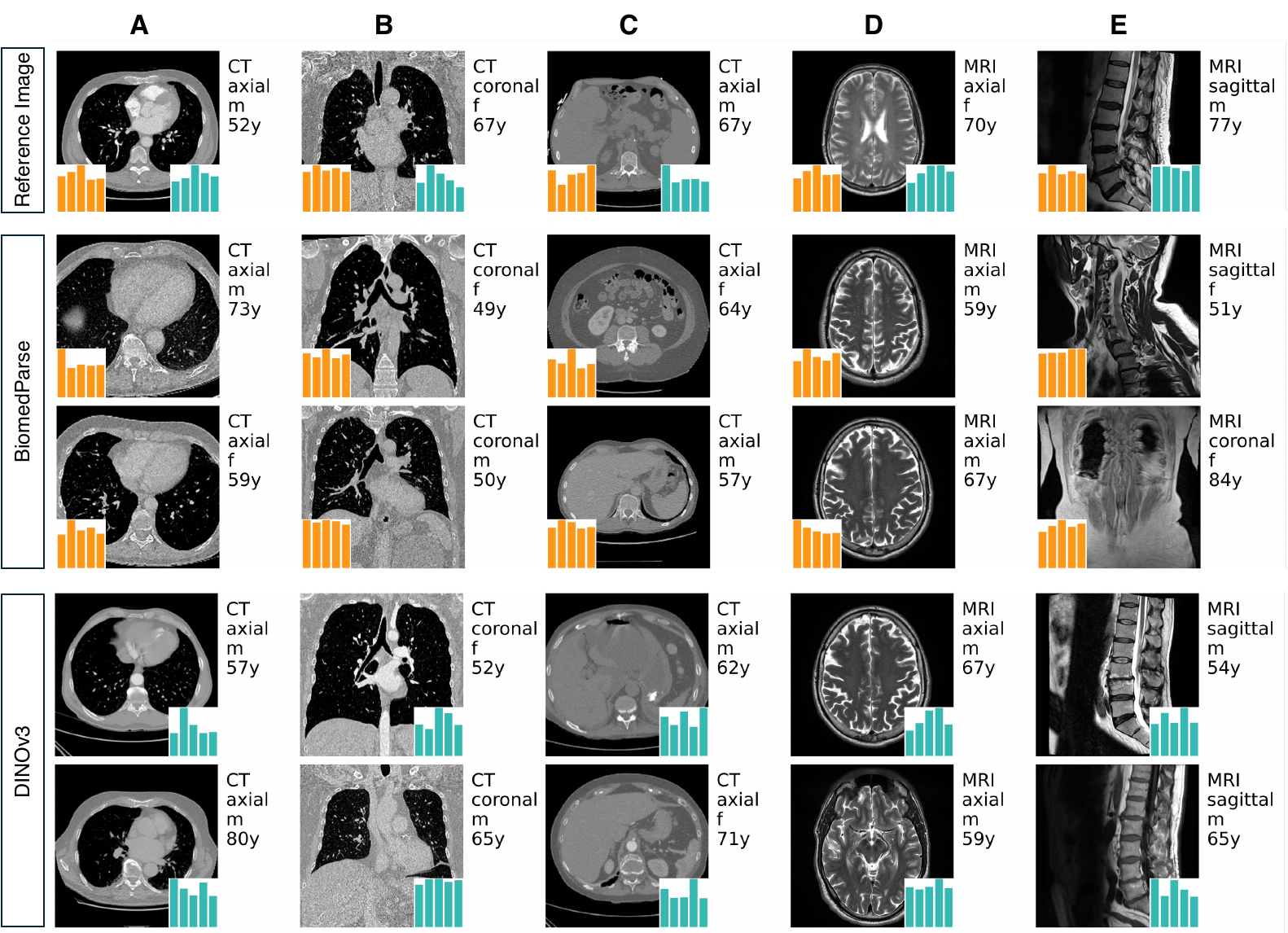}
    \caption{Sparse fingerprint retrieval at $k=5$ for five reference cases (A--E) spanning CT and MRI across multiple anatomical regions.
    Row 1: reference images with BiomedParse (orange) and DINOv3 (blue) fingerprint insets.
    Rows 2--3: top-2 BiomedParse retrievals. Rows 4--5: top-2 DINOv3 retrievals.}
    \label{fig:sparse-fingerprint}
\end{figure}

\subsection{Sparse Feature Interpretability}
\label{subsec:sparse-feature-interpretability}

For the optimal configurations per FM (Table \ref{tab:final-ranking}), we evaluate sparse feature interpretability through three demonstrations, excluding the random-weight baseline due to its lack of semantic structure (Sect.~\ref{subsec:sae-quality}).

\paragraph{Sparse feature-based image retrieval.}

We evaluate sparse fingerprints, the top-$k$ active features per image, for image retrieval to assess whether sparse features preserve semantic similarity.
Retrieval quality is measured as mean cosine similarity of the top-5 retrieved images for 1,000 randomly selected reference images of the test set in the dense embedding space (Table \ref{tab:sparse-fingerprint}).
At $k=5$ features (Fig.~\ref{fig:sparse-fingerprint}), BiomedParse achieves 97.7\,\% of dense retrieval quality (0.954 vs. 0.976) and DINOv3 achieves 92.8\,\% (0.831 vs. 0.895).
Quality saturates rapidly above $k=10$ for both models, confirming that semantic content concentrates in a small number of sparse features.

\begin{table}[t]
    \centering
    \caption{Sparse fingerprint retrieval quality (mean cosine similarity to the reference image in the dense embedding space) as a function of fingerprint size $k$, averaged over $N=1{,}000$ test images. \emph{Dense}: full dense retrieval quality (upper bound).}
    \label{tab:sparse-fingerprint}
    {\scriptsize
    \setlength{\tabcolsep}{6pt}
    \begin{tabular}{lccccc}
    \hline
    \textbf{Model} & \textbf{k=1} & \textbf{k=5} & \textbf{k=10} & \textbf{k=20} & \textbf{Dense} \\
    \hline
    BiomedParse & 0.93 & 0.95 & 0.96 & 0.97 & 0.98 \\
    DINOv3 & 0.75 & 0.83 & 0.85 & 0.86 & 0.89 \\
    \hline
    \end{tabular}}

    \medskip

    \caption{LLM-as-judge evaluation of automatically generated feature concepts for $N=100$ features per model.
    An independent VLM ranks the true concept description among five candidates (1 true $+$ 4 distractors) given the same images.
    Rank 1 = true concept fits best, Rank 5 = true concept fits worst.
}
    \label{tab:autointerp-results}
    {\scriptsize
    \setlength{\tabcolsep}{6pt}
    \begin{tabular}{lcccccc}
    \hline
    \textbf{Model} & \textbf{Mean rank} & \textbf{Rank 1} & \textbf{Rank 2} & \textbf{Rank 3} & \textbf{Rank 4} & \textbf{Rank 5} \\
    \hline
    BiomedParse & 1.48 & 70 & 17 & 9 & 3 & 1 \\
    DINOv3 & 1.26 & 83 & 11 & 3 & 3 & 0 \\
    \hline
    \end{tabular}}
\end{table}

\paragraph{Interpretable sparse feature concepts.}

We interpret the top-100 interpretable features per model by automated VLM-based concept generation, verified by an independent LLM judge that ranks the true description among five candidates (rank 1 = best, rank 5 = worst).
DINOv3 achieves 83/100 rank-1 counts (mean rank 1.26), outperforming BiomedParse (70/100 rank-1 counts, mean rank 1.48).
Rank-2 counts are 11/100 and 17/100, respectively (Table \ref{tab:autointerp-results}).
Concepts capture modality, imaging plane, anatomy, and demographics, emerging from self-supervised learning without explicit anatomical labels.

\paragraph{Language-based image retrieval.}

\begin{figure}[t]
\centering
\includegraphics[width=0.8\textwidth]{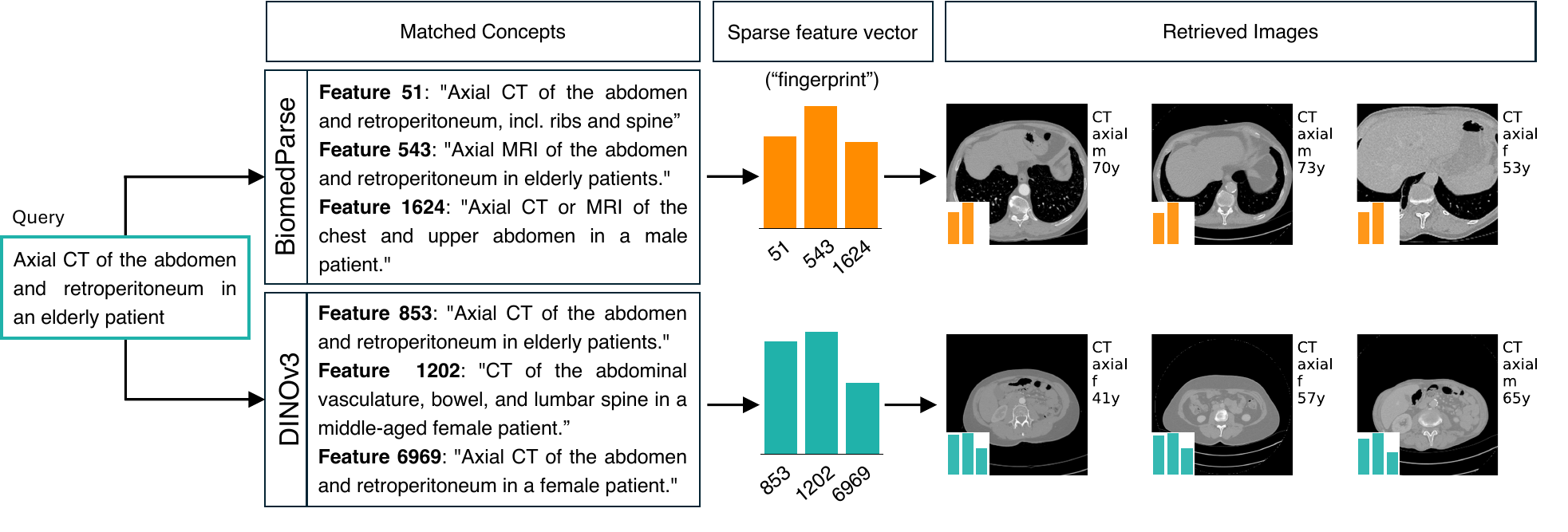}
\caption{Zero-shot language-driven retrieval for ``Axial CT of the abdomen and retroperitoneum in an elderly patient.''
An LLM selects matching feature concepts (left), determining a sparse fingerprint (center) for cosine retrieval (right).
BiomedParse selects mixed MRI/CT concepts and retrieves thoracic images.
DINOv3 selects CT-specific abdomen features and retrieves correct axial abdominal CT.}
\label{fig:language-based-image-retrieval}
\end{figure}

As an end-to-end demonstration, an LLM maps a clinical text query to matching sparse feature concepts and assembles a sparse fingerprint for cosine retrieval, requiring no reference image or task-specific training.
For the query ``Axial CT of the abdomen and retroperitoneum in an elderly patient'' (Fig.~\ref{fig:language-based-image-retrieval}), BiomedParse's features lacks a modality-pure abdomen feature, selects mixed MRI/CT concepts, and retrieves thoracic images.
DINOv3, whose richer feature vocabulary includes three anatomy- and modality-specific abdomen CT concepts, retrieves anatomically correct axial abdominal CT images.
This shows that automatically labeled, unsupervised concepts can bridge clinical language and medical image content, capturing anatomy and modality while demographic constraints remain open, complementing text-supervised retrieval such as BiomedCLIP \cite{zhang2025b} or PubMedCLIP \cite{eslami2023findings} without requiring paired image-text supervision.

\section{Conclusion}

Sparse features from Matryoshka SAEs faithfully preserve embedding structure, recover strong downstream performance with a handful of features, and enable interpretable retrieval and zero-shot language-driven search, extending prior evidence from chest radiographs to multi-modal volumetric imaging across two architecturally distinct foundation models.

DINOv3, despite no biomedical pretraining focus, consistently produces more monosemantic features and comparable downstream performance, suggesting that representational richness matters more than domain alignment for interpretability.
Language-driven retrieval, a single-query proof-of-concept, captures anatomy and modality, with finer-grained constraints such as demographics an open direction.
Monosemanticity scoring relies on metadata-derived organ labels and VLM-generated concept descriptions rather than human annotation, providing scalable but proxy-based evidence, with expert evaluation a key next step.
The TotalSegmentator dataset covers normal anatomy across 10 institutions and two modalities but excludes pathological cases, and analysis operates at the 2D slice level rather than volumetrically.
Because concepts are described from model-specific images, concept quality tracks sparse-feature quality, leaving cross-FM concept similarity, broader clinician-validated query benchmarks, and comparison with text-supervised retrieval as future work.

Overall, sparse autoencoders provide a practical interpretability layer for self-supervised medical vision models, requiring no architectural modification, task-specific labels, or retraining.
By bridging abstract FM representations and human-interpretable concepts, sparse autoencoders offer a grounded path toward medical AI systems whose predictions can be inspected, communicated, and trusted in clinical practice.

\begin{credits}
\subsubsection{\ackname}
\label{subsubsec:acknowledgments}

Philipp Wesp is funded by the Deutsche Forschungsgemeinschaft (DFG, German Research Foundation) - 553239084.

\subsubsection{\discintname}
\label{subsubsec:disclosure-of-interests}
The authors have no competing interests to declare that are relevant to the content of this article.
\end{credits}

\newpage

%
%
%
\bibliographystyle{splncs04}
\bibliography{mybibliography}
\end{document}